\documentclass[11pt,a4paper]{article}
\usepackage{times}
\usepackage{latexsym}
\usepackage{comment}
\usepackage[hyperref]{acl2020}

\usepackage{microtype}

\aclfinalcopy 
\def\aclpaperid{60} 

\usepackage{amsmath}
\usepackage{amssymb}
\usepackage{booktabs}
\usepackage{mathrsfs}
\usepackage{bbm}
\usepackage{graphicx}
\usepackage{multirow}

\title{
Beyond User Self-Reported Likert Scale Ratings: \\
A Comparison Model for Automatic Dialog Evaluation  
}

\author{Weixin Liang$^{1}$, 
James Zou$^{1}$ and Zhou Yu$^2$\\ 
~~\\$^1$ Stanford University,$^2$ University of California, Davis
~~\\ \{wxliang, jamesz\}@stanford.edu, joyu@ucdavis.edu
}

\date{}

\newcommand{\modelname}[1]{{{{Comparison Model for Automatic Dialog Evaluation}}}{#1}}
\newcommand{\modelabbrevname}[1]{{{{CMADE}}}{#1}}

\begin{document}
\maketitle
\begin{abstract}

Open Domain dialog system evaluation is one of the most important challenges in dialog research. 
Existing automatic evaluation metrics, such as BLEU are mostly reference-based. They calculate the difference between the generated response and 
a limited number of available references. 
Likert-score based self-reported user rating is widely adopted by social conversational systems, such as Amazon Alexa Prize chatbots. However, self-reported user rating suffers from bias and variance among different users. To alleviate this problem, we formulate dialog evaluation as a comparison task. We also propose an automatic evaluation model \modelabbrevname{} (\modelname) that automatically cleans self-reported user ratings as it trains on them.  Specifically, we first use a self-supervised method to learn better dialog feature representation, and then use KNN and Shapley to remove confusing samples. 
Our experiments show that \modelabbrevname{} 
 achieves 89.2\% accuracy in the dialog comparison task. 
Our implementation is available at  \url{https://github.com/Weixin-Liang/dialog_evaluation_CMADE}.

\end{abstract}

\section{Introduction}

Open-domain dialog system evaluation is one of the most difficult challenges in the dialog community. 
Open-domain chatbots have a user-centric goal: to provide human with enjoyable user experience. 
However, user experience is difficult to quantify due to  bias and variance among different users. Previous research has optimized on automatic 
dialog evaluation metrics such as BLUE, which 
measures the difference between the generated responses and the reference responses. 
Due to the 
contrast between the 
one-to-many nature of open-domain conversations and the limited number of available references, 
such metrics correlate poorly with human judgments~\cite{howNotEval}. 
Designing a fully automatic dialog evaluation metric is still an open research problem.

Currently, 
both academia and industry~\cite{li2019end,MOSS} rely on 
human ratings to evaluate open-domain dialogs. 
Following the ubiquitous application of Likert scores in 
survey research like online reviews~\cite{marketing1} and consumer satisfaction~\cite{marketing1992,DBLP:conf/lak/WangLHO19}, 
a common practice of human evaluation on dialogs is to ask 
either a third-person rater or the 
chatbot user  
to report a Likert score. 
However, concerns have been raised about 
the validity of Likert score-based ratings. 
Kulikov et al.~\cite{DBLP:journals/corr/abs-1811-00907} observe high bias and variance of Likert scores. 
Such issue is more severe in real-world commercial dialog systems like Alexa social chatbot~\cite{alexaEval},
because the real-world users 
have neither monetary incentive nor necessary annotation training 
to calibrate their ratings.

To explore the validity of Likert score based dialog evaluation, 
we first perform a large-scale data analysis of 
3,608 collected real-world human-machine dialogs along with their self-reported Likert scale ratings from Amazon Alexa Prize Challenge~\cite{gunrock,chen2018gunrock}. 
One noticeable property of the ratings is its J-shape skew distribution: nearly half of the dialogs are rated with the highest Likert score. 
The prevalence of such extreme distribution of ratings 
has long been observed by the business research community in variable aspects of real-life~\cite{marketing25platforms,marketing1,marketing2}.

Although we could tell which dialog system is better by running statistical test on a large number of noisy ratings, 
it is difficult to locate dialogs with bad performance reliably to improve dialog system quality. 
In this paper, 
we take on the challenge of 
calibrating a large number of noisy self-reported user ratings to build better dialog evaluation models. 
We formulate the task as to first
denoise the self-reported user ratings and then train a model on the cleaned ratings. 
We design \modelabbrevname{} (\modelname), 
a progressive three-stage 
denoising pipeline. 
We first perform a self-supervised 
learning to obtain good dialog representations. 
We then fine-tune
\modelabbrevname{} on smoothed self-reported user ratings to improve the dialog representation while preventing the network from overfitting on noisy ratings. 
Finally, we apply data Shapley 
to remove noisy training data, and fine-tune the model on the cleaned training set. 
Our experiments show that 
\modelabbrevname{} is able to successfully identify noisy training data and achieves 89.2\% in accuracy and 0.787 in Kappa on a test set with unseen expert-rated dialog pairs.


\section{Related Work}

Open-domain dialog system evaluation is a long-lasting challenge. 
It has been shown that 
previous automatic dialog evaluation metrics correlate poorly with human judgments~\cite{howNotEval,ademACL,whyWeNeedNewEval}. 
A well-known reason is that 
these automatic dialog evaluation metrics 
rely on modeling the distance between the generated response and 
a limited number of references available. 
The fundamental gap between the open-ended nature of the conversations 
and the limited references~\cite{multiRefTianchengZhao} 
is not addressed in methods that are lexical-level based~\cite{bleu,rouge,meteor}, 
embedding based~\cite{greedyMatching,forgues2014bootstrapping}, 
or learning based~\cite{ruber,ademACL}.

Given the aforementioned limitations, 
Likert-score based rating is the de-facto standard 
for current dialog research and 
social conversational systems such as in Amazon Alexa Prize Challenge~\cite{gunrock,chen2018gunrock}. 
Various forms of evaluation settings have been explored to better measure human judgments. 
Single-turn pairwise  comparison~\cite{singleTurnAB1,singleTurnAB2} 
is primarily used for comparing two dialog systems. 
Each system predicts a single utterance 
given the static ``gold'' context utterance 
from human-human logs. 
Although such A/B test setting is robust to annotator score bias, 
it cannot capture the multi-turn nature of dialogs. 
A more complete multi-turn evaluation is typically
measured with a Likert scale 
for the full dialog history, where 
either a third-person rater or the chatbot user~\cite{WhatGoodConversation} reports a Likert score on 
user experience~\cite{alexaEval}, 
engagement~\cite{engagement}
or appropriateness~\cite{ademACL}. 
However, as observed in ~\cite{DBLP:journals/corr/abs-1811-00907,alexa,alexaEval} 
Likert scores suffer from bias and variance among different users. 
Different from previous empirical observations, 
we conduct a large-scale quantitative and qualitative data analysis of Likert score based ratings. 
To address the issue of Likert scores, 
the Alexa team proposed a rule-based ensemble of turn-granularity expert ratings~\cite{AlexaEvalDilek}, and automatic metrics like topical diversity~\cite{AlexaTopicEval} and conversational breadth. 
ACUTE-EVAL~\cite{ACUTEEVAL} makes a small-scale attempt to use multi-turn pair-wise comparison to rank different chatbots. 
Given the ubiquity and simplicity of Likert scores based evaluation, 
instead of proposing an alternative measure, 
we take on the challenge of denoising Likert scores 
with minimal expert annotations introduced (one order of magnitude smaller). 
Different from~\cite{ACUTEEVAL}, 
our proposed expert annotation scheme is for comparing 
the dialogs within the same chatbot.

\begin{figure*}[h!tb]
\centering
\includegraphics[width=\textwidth]
{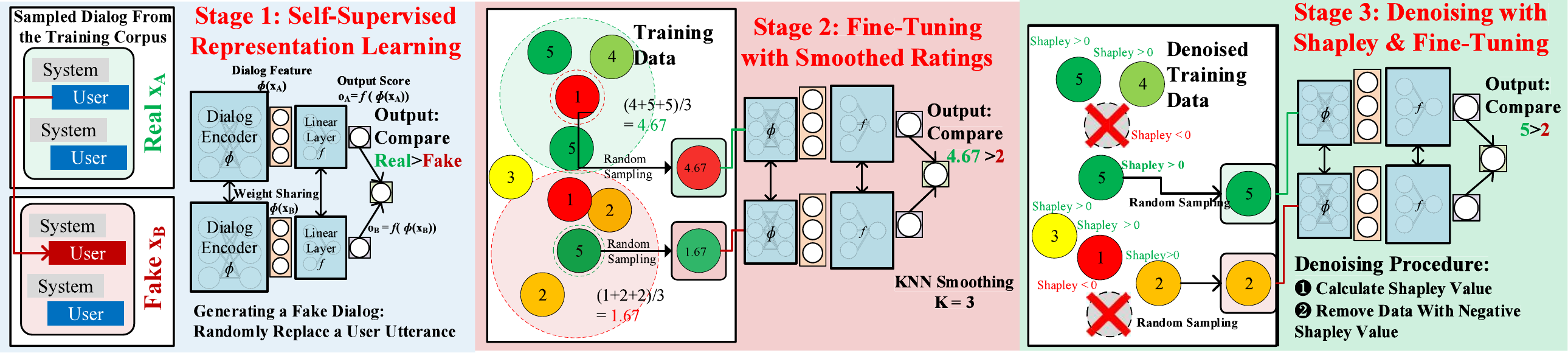} 
\caption{
Schematic of the \modelabbrevname{} workflow. 
\modelabbrevname{} contains a three-stage training pipeline to denoise self-reported ratings to train an automatic dialog comparison model: 
learning representation viaself-supervised dialog flow anomaly detection, 
fine-tuning with smoothed self-reported user ratings,
denoising with data Shapley \& further fine-tuning. 
The gray and blue rectangles in stage 1 represents system and user utterances. The red rectangle in stage 1 represents the randomly replaced system utterance 
for dialog flow perturbation. 
In stage 2 \& 3, each ball represents a dialog in the training data. The number on each ball represents the dialog rating. 
}
\label{fig:main}
\end{figure*}

\section{Dialog Rating Study}
\label{sec:study}

\subsection{Dataset}
The data used in this study was collected 
during the 2018 Amazon Alexa Prize Competition~\cite{alexa2018}. 
Our data contain 
long and engaging spoken conversations between thousands of real-world Amazon Alexa customers and 
Gunrock, the 2018 Alexa Prize winning social bot~\cite{gunrock}.
The chatbot has 11 topic dialog modules including movies, books, and animals. 
One notable characteristic of the chatbot is its versatile and complex dialog flows which interleaves 
facts, opinions and questions 
to make the conversation flexible and interesting~\cite{chen2018gunrock}.  
At the end of each dialog, a self-reported Likert scale rating is elicited by the question ``on a scale of one to five, how likely would you talk to this social bot again?''

We first filter out dialogs that have inappropriate content using keyword matching. We then select 3,608 ten-turn dialogs on movies,  because movie dialogs are more coherent and diverse compared to other topics according to both real users and Amazon selected experts. 
We observe that dialogs with more than eight turns are more meaningful and semantically versatile, while dialogs more than 10 turns  
exceed the max length limit of the BERT model (512 tokens). So we select dialogs that have ten turns.
Our approach could support longer conversations by adopting a memory footprint efficient algorithm for self-attention to support sequences with thousands of tokens~\cite{musicTransformer}. We leave this to future work.

We aim to evaluate user experience for each dialog from the same chatbot of the same length. This is significantly more challenging than identifying which chatbot provides a better user experience on average 
since our problem setup requires us to capture more subtle difference in user experience.

\subsection{Likert Score Based Evaluation}
\label{subsec:Likert}

\begin{table}[htb]
\footnotesize
\begin{center}
\begin{tabular}{rlllll}
\cmidrule[\heavyrulewidth]{1-6}
\textbf{Rating} & \textbf{1}      & \textbf{2}      & \textbf{3}      & \textbf{4}      & \textbf{5}      \\ 
\cmidrule{1-6}
Count  & 386    & 404    & 566    & 664   & 1588   \\ 
\cmidrule{1-6}
Fraction  & 10.7\% & 11.2\% & 15.7\% & 18.4\% & 44.0\% \\ 
\cmidrule[\heavyrulewidth]{1-6}
\end{tabular}
\caption{
The statistics of self-reported Likert scale ratings. 
The distribution is heavily skewed and noisy: nearly half of the dialogs are rated with  score = 5. 
}
\label{tab:userRatingStat}
\end{center}
\end{table}

\noindent \textbf{J-Shape Skewness}
We perform a detailed analysis of the self-reported Likert scale ratings.
As shown in Table~\ref{tab:userRatingStat}, 
abnormally, nearly half of the dialogs are rated as five, which is the highest score. 
A similar skewed distribution is also observed in previous years' Alexa competition~\cite{UW2017alexaWin}.  
In fact, the business research community has long observed the prevalence of the extreme distribution of reviews in which the reviews are heavily skewed to the positive end of the rating scale (known as "J-shape") 
in online reviews (e.g., Amazon, Airbnb, Yelp)~\cite{marketing1,marketing2,marketing3}, 
word of mouth~\cite{marketing4} 
and consumer satisfaction~\cite{marketing1992,marketing1996}. 

\noindent \textbf{Comparison to expert ratings}
We randomly selected $50$ dialogs rated score-5 and showed these to an expert, and our expert rated $27$ of them with score-4 or less. 
The Alexa team~\cite{alexaEval} has also reported that the inter-user agreement is quite low for their internal rating analysis.
Such phenomena indicate that the self-reported Likert scale ratings are extremely noisy. Using such ratings cannot localize individual bad interactions. 
In addition, Likert score based evaluation also suffers from insensitivity issues. 
As observed by the Alexa team~\cite{alexaEval} in 
multiple 
internal user studies, 
even though users evaluated multiple dialogs with the same score, they had a clear rank order among the dialogs.

The skewness, noisiness and insensitivity of the self-reported Likert scale rating make it  a sub-optimal dialog evaluation metric. 
In practice, we find that directly training a classifier (even for pre-trained BERT-based model) 
on the noisy self-reported Likert scale ratings 
suffers from underfitting. 
One of the Alexa Price Challenge team, 
Alana~\cite{AlexaTeam2017alana} train a binary-classifier between 
successful dialogs (human rating 4 or 5) and unsuccessful dialogs (rating 1 or 2)
with heavy hand-engineered features. 
They reach 69.40\% accuracy on this binary classification problem, which is far from usable in real-world settings.

\subsection{Pairwise Comparison Based Evaluation}
\label{subsec:pairwise}
Selecting the better dialog from two options is easier for a human evaluator than giving an absolute number like the Likert score, which requires the evaluator to maintain a consistent standard. 
People's perception is inherently relative, and pair-wise comparison is local and does not require the user to have global consistency. 
There are many other examples where
humans find it easier to perform pairwise comparisons rather than providing direct labels~\cite{reviewer3,deepstore,memsec}, including
content search~\cite{comparisonContentSearch}, 
image retrieval~\cite{comparisonImageSearch,DBLP:conf/isbi/ZhaoLGZML19,ijcv}, 
and age estimation~\cite{ageEstimationPairPosterior}.

We randomly sample 400 
dialog pairs for experts to annotate. 
We ask the question, 
``If you were the user, in which scenario would you be more likely to come back and talk to the system again? '' 
We guide the experts to focus on the user experience rather than calibrating the performance of any specific module of the dialog system. 
Two researchers with conversational training experience annotated the data. The leading expert has been working in an Alexa competition team for more than one year with an emphasis on the user ratings. For each dialog pair $(A, B)$, they label `$A$ is better than $B$' or `$B$ is better than $A$' or `cannot tell'. They reached a high inter-annotator agreement score~\cite{cohen1968weightedkappa} with kappa $ \kappa = 0.83 $. To make sure that the dev \& test is accurate, we throw away all ``cannot tell'' dialog pairs. 
We then study the correlation between 
Likert score based evaluation
and 
pairwise comparison based evaluation.

\subsection{Correlation Between User Ratings and Expert Ratings}
\label{subsec:correlation}
\begin{table}[htb]
\footnotesize
\begin{center}
\begin{tabular}{rllll}
\cmidrule[\heavyrulewidth]{1-5}
\begin{tabular}[c]{@{}c@{}}Delta of Self-Reported \\ Ratings (e.g., 5-1=4)\end{tabular} & \textbf{$\Delta$=1}      & \textbf{$\Delta$=2}      & \textbf{$\Delta$=3}      & \textbf{$\Delta$=4} \\ 
\cmidrule{1-5}
Disagreement Rate  & 0.45 & 0.383 & 0.220 & 0.157   \\ 
\cmidrule[\heavyrulewidth]{1-5}
\end{tabular}
\caption{
The correlation between the self-reported Likert scale ratings and our pair-wise comparison annotation. 
For a pair of dialogs, if the delta of self-reported Likert scale ratings is large, then they are more likely to
align with the comparison results from experts. 
}
\label{tab:disagreement}
\end{center}
\end{table}

To further analyze the self-reported Likert scale ratings, 
we also compare the annotated labels of the 
403 dialog pairs 
with the self-reported Likert scale ratings of these dialogs. 
For each pair of dialogs, we compare 
the pairwise comparison label and the delta between the self-reported Likert scale ratings of the two dialogs. 
Ideally, the dialog with a higher self-reported Likert scale rating should be the one that is annotated as having a better user experience in the pairwise comparison. 
We count the number and fraction of ``disagreement'' between the two types of ratings. 
Overall, roughly 1/3 of the dialog pairs disagree. 
As shown in Table~\ref{tab:disagreement}, 
as the gap between the self-reported Likert scale ratings becomes larger, the disagreement between expert and self-reported ratings goes down. 
This suggests that if the difference between the two dialogs' Likert score is huge, they are more likely to be consistent with the comparison ratings.

\section{Problem Formulation} 
Suppose the training set $D_{train}$ consists of data points 
    $ D_{train} = \{(x_i, y_i)\}^{N_{train}}_1 $ 
    where $x_i$ is a dialog and $y_i$ is the noisy self-reported user ratings. 
    We define a 
    strict partial order
    relationship $\triangleright$ where 
    $x_i \triangleright x_j $ means that 
    dialog $x_i$ provides a better user experience than dialog $x_j$. 
    Note that $y_i > y_j$ does not always imply $x_i \triangleright x_j $ since self-reported user ratings are noisy (\S~\ref{subsec:pairwise}, \S~\ref{subsec:correlation}).
    The test set $D_{test}$ consists of $N_{test}$ dialog pairs along with their binary pair-wise comparison labels 
    $ D_{test} = \{ (x^{test}_{i}, x^{test}_{j}, z^{test}_{i,j}) \}_{i,j \in I^{test}} 
    $,
    where $z^{test}_{i,j}$ is annotated by experts and indicates whether dialog A provides a better user experience than dialog B, i.e., $ z^{test}_{i,j} = \mathbbm{1}(x_i \triangleright x_j)$. 
    The development set $D_{dev}$ has a similar structure. 
    
    Following the structure of the expert annotated pairs, 
    we formulate our model $M(\phi, f)$ as a pair-wise dialog predictor with a similar architecture as RankNet~\cite{ranknet}. 
    For a dialog pair $(x_i, x_j)$, 
    the model predicts an un-normalized score $ o_i, o_j \in \mathbb{R}$ for each dialog:
        $o_i  = f(\phi(x_i))$ and  
        $o_i = f(\phi(x_j))$
    where $\phi$ is a dialog encoder that maps each dialog to 
    a feature space 
    and $f$ is a linear transformation that converts each dialog feature into a real number $o$. 
    We define a binary relationship $\hat{\triangleright} $ where 
    $x_i \hat{\triangleright} x_j$ means that the model predicts that dialog $x_i$ provides a better user experience than dialog $x_j$. 
    We denote model's prediction of $z_{i,j}$ as $\hat{z_{i,j}}$ where $\hat{z_{i,j}}=\mathbbm{1}(x_i \hat{\triangleright} x_j)$. 
    We model the predicted posterior $ P(\hat{z_{i,j}}=1) = P(x_i \hat{\triangleright} x_j)$ as: 
    \begin{equation*}
        P(\hat{z_{i,j}}=1) = P(x_i \hat{\triangleright} x_j) 
= \frac{1}{ 1+e^{-(o_i - o_j)} }
    \end{equation*}
    
\section{Method}
\label{sec:method}   
    Our goal is to reduce the noise of the self-reported user ratings~(\S~\ref{sec:study}). 
    Directly training a classification 
    model using the noisy ratings leads to severe underfitting. 
    To this end, 
    we propose a three-stage training pipeline to 
    denoise self-reported ratings to  train an automatic dialog comparison model.
     Figure~\ref{fig:main} describes the overall pipeline: 
    \begin{itemize}
        \item In Stage 1, 
        we learn
        dialog feature representation  with a self-supervised dialog flow anomaly detection task. 

        \item In Stage 2, we perform label smoothing to adjust the noisy self-reported ratings in the training set and fine-tune the dialog comparison model on the smoothed ratings. 
        
        \item In Stage 3, we perform data Shapley~\cite{jamesShapley,boxinShapley} on the self-reported user ratings to identify and remove noisy data points. We then fine-tune the dialog comparison model on the cleaned training set. 
    \end{itemize}

\subsection{Stage 1: Learning Representation via self-supervised dialog  anomaly detection } 
\begin{table}[ht]
\footnotesize
\centering
\begin{tabular}{p{7cm}}
\cmidrule[\heavyrulewidth]{1-1}
\textbf{Sys:}  What movie did you see?  \\
\textbf{User:}  Spider man into the spider verse  \\
\cmidrule{1-1}
\textbf{Sys:}  Ah, I know about Spider man into the spider verse! I'm wondering. What would you rate this movie on a scale from 1 to 10? \\ 
\cmidrule{1-1}
\textbf{Replaced Sys:}  Isn't it crazy how famous actors can get? 
Are you interested in talking more about Scarlett Johansson? \\
\cmidrule[\heavyrulewidth]{1-1}
\end{tabular}
\caption{
A fake dialog example created by dialog flow perturbation in Stage 1. 
We perturb the dialog flow by 
replacing a system utterance (here the second Sys utterance in the table) with a random system utterance from the corpus (here the replaces Sys utterance) to generate a fake dialog. 
With high probability, the fake dialog is less appropriate than the origin one. 
}
\label{table:exampleDialog}
\end{table}

Having a good dialog representation is the first step towards denoising the data.  
Our primary goal in this stage is to 
train a dialog encoder $\phi$ to learn good dialog feature representations for the following stages. 
Here $\phi$ could be any sequence encoder that could encode a dialog 
and we use BERT~\cite{bert} in this paper.

For each dialog in the training set, 
we perturb the dialog flow to generate a fake dialog 
and train the model to 
differentiate the fake dialog and the real one. 
Dialog flow is a user-centric measure of whether a conversation is ``going smoothly''~\cite{beyondTuring}. 
To perturb the dialog flow for each dialog $x_i$, 
we randomly replace a user utterance in $x_i$ with a random user utterance from the training corpus $D_{train}$, 
yielding a perturbed dialog $x_{i,fake}$. 
With high probability, 
the system utterance immediately following the replaced user utterance becomes inappropriate. 
Therefore, we incorporate 
$ \{(x_i, x_{i,fake}, z=1)\} $ into the training pairs. 
Similarly, we also randomly replace a system  utterance and yield another perturbed dialog. 
We generate two perturbed dialogs 
for each dialog in the training set and thus $2N_{train}$ real-fake dialog pairs in total. 
An example is shown in Table~\ref{table:exampleDialog}. 
We note that appropriateness is one of the most widely applied metrics of human evaluation on dialogs~\cite{ademACL}. 
By learning to differentiate the perturbed dialog and the original one, 
we expect 
\modelabbrevname{} to learn a good dialog encoder $\phi$ 
which maps dialogs with similar dialog flow close to each other in the feature space.

\subsection{Stage 2: Fine-tuning with smoothed self-reported user ratings} 
Stage 1 only performs self-supervised learning~\cite{yu2020steam,chen2020semi,DBLP:journals/corr/abs-2005-10636} 
and does not incorporate any supervision from human ratings. 
To obtain better dialog feature representations for Stage 3, 
Stage 2 fine-tunes $\phi$ with supervision from the noisy self-reported user ratings. 
We adopt a simple yet effective label smoothing, inspired by~\cite{labelSmooth,allen}, 
using 
the representation learned 
in 
Stage 1. 
A key assumption in Stage 2 
is that 
dialogs with similar dialog flow provide a similar user experience. 
For each dialog $x_i$, 
we find its $K$ nearest neighbors in the feature space defined by $\phi$. 
We use the average self-reported ratings of the $K$ nearest neighbors as a smoothed rating $y_i^s$ for $x_i$. 
To construct training dialog pairs, 
we randomly sample dialog pairs $x_i$ and $x_j$ and derive 
a pair-wise comparison label 
$ z^s_{i,j} $ by comparing the smoothed rating 
${y_i^s}$ and ${y_j^s}$:
$ z_{i,j}^s =  \mathbbm{1}({y_i^s} > {y_j^s}).$ 
We discard the pairs with equal 
$y_i^s$ and $y_j^s$. 
To improve the dialog feature representation, 
we fine-tune the model $M(\phi, f)$ on sampled dialog pairs along with the derived labels from comparing the smoothed scores
$ \{ x_i, x_j, z^s_{i,j} \} $. 
We note that $z^s_{i,j}$ depends solely on the noisy self-reported ratings in the training set 
and does not depend on the expert annotations. 
Theoretically, 
we could iterate between label smoothing and model fine-tuning 
since the fine-tuned model provides better dialog feature representation. 
In practice, we find that one iteration is enough 
to reach good prediction performance.

Label smoothing has led to state-of-the-art models 
in image classification~\cite{labelSmooth}, language translation~\cite{attentionIsAllYouNeed} and 
speech recognition~\cite{speechLabelSmoothApplication}.
Prior attempts in label smoothing~\cite{labelSmooth,attentionIsAllYouNeed,speechLabelSmoothApplication,labelSmoothWhen} 
focus on categorical labels to 
prevent the network from becoming over-confident 
while 
we apply label smoothing on ordinal labels (i.e., Likert scores)
to 
prevent the network from 
overfitting on noisy ordinal labels.

\subsection{Stage 3: 
Denoising with 
data Shapley \& further fine-tuning}
\label{subsec:shapley}
In Stage 2, noisy ratings still have effect in the smoothed ratings for other data points. 
In Stage 3, 
we aim to identify and remove 
dialogs with noisy self-reported user ratings $y_i$ 
with data Shapley value  technique~\cite{jamesShapley,boxinShapley, boxinShapley2}. 
Shapley value comes originally from cooperative game
theory~\cite{dubey1975uniqueness}. 
In a cooperative game, 
there are $n$ players $D = \{ 1,...,n \}$ and 
a utility function $v: 2^{[n]} \rightarrow  \mathbb{R} $
assigns a reward to each of $2^n$ 
subsets of players: 
$v(S)$ is the reward if the players in subset $S \subseteq D$ cooperate. 
Shapley value defines 
a unique scheme to distribute the total gains generated by the
coalition of all players $v(D)$ with 
a set of appealing mathematical properties. 
Shapley value has been applied to problems in various domains, ranging from economics~\cite{shapleyEconomics}
to machine learning~\cite{2005featureSelectionShapley,yona2019s}.

In our setting, 
given $ D_{train} = \{(x_i, y_i)\}^{N_{train}}_1 $, 
we view them as $N_{train}$ players. 
We could also view the utility function $v(S)$ 
as the performance on the development set. 
The Shapley value for player $i$ 
is defined as the average marginal 
contribution of $\{(x_i, y_i)\}$ to all possible subsets that are 
formed by other users~\cite{boxinShapley}:

    $$ s_{i} = \frac{1}{N} \sum_{S \subseteq D_{train} \setminus \{x_i\} } 
    \frac{1}{\binom{N-1}{|S|}} 
    [v(S \cup \{x_i\}) - v(S)]
    $$

As suggested by the definition of data Shapley, 
computing data Shapley value requires an exponentially large number of computations
to enumerate $\mathcal{O}(2^{N_{train}})$ 
possible subsets and train the model $M$ on each subset, which is intractable. 
Inspired by~\cite{boxinShapley}, 
\modelabbrevname{} tackles this issue by 
reducing the deep model $M$ to a 
k-nearest neighbors (KNN) model 
and then apply 
the closed-form solution 
of shapley value on KNN. 
Using the feature extractor $\phi$ trained in Stage 1 and Stage 2, 
we fix $\phi$ and map all dialogs in the training data $\{x_i\}_1^{N_{train}}$ 
to $\{\phi(x_i)\}_1^{N_{train}}$. 
We first define the utility function $v(S)$ in a special case 
where the development set only contains one dialog pair $(x^{dev}_{p}, x^{dev}_{q}, z^{dev}_{p,q})_{p,q \in I^{dev} = \{ (1,2) \} }$. 
In our setting, the development set contains dialog pairs annotated by experts. 
Given any nonempty subset $S \subseteq D_{train}$, 
we use the KNN Regressor to rate $x^{dev}_{p}$ and $x^{dev}_{q}$. 
To do this, 
we compute $ \phi(x^{dev}_{p}) $ and sort $\{x_p\}_1^{N_{train}}$ 
based on their euclidean distance in the dialog feature space to $x^{dev}_{p}$, 
yielding 
    $ (x_{\alpha^{(p)}_{1}}, x_{\alpha^{(p)}_{2}}, ..., x_{\alpha^{(p)}_{|S|}} )$
with $x_{\alpha^{(p)}_{1}}, ..., x_{\alpha^{(p)}_{K}}$ 
as the top-K most similar dialogs to  $x^{dev}_{p}$. 
Similarly, we get 
    $ (x_{\alpha^{(q)}_{1}}, x_{\alpha^{(q)}_{2}}, ..., x_{\alpha^{(q)}_{|S|}} )$
 with $x_{\alpha^{(q)}_{1}}, ..., x_{\alpha^{(q)}_{K}}$ 
as the top-K most similar dialogs to  $x^{dev}_{q}$. 
Based on the self-reported user ratings in the training data, 
we use the KNN Regressor to rate $x^{dev}_{p}$ and $x^{dev}_{q}$ as follows: 
\begin{align}
    \hat{y}^{dev}_{p} & = \frac{1}{K} \sum_{k=1}^{min
    \{K,|S|\}} y_{\alpha^{(p)}_{k}} \\ 
    \hat{y}^{dev}_{q} & = \frac{1}{K} \sum_{k=1}^{min
    \{K,|S|\}} y_{\alpha^{(q)}_{k}}
\end{align}
The model predicts $ \hat{z}^{dev}_{p,q} = 1 $ if 
$\hat{y}^{dev}_{p} > \hat{y}^{dev}_{q}$ and vice versa.

To obtain a closed-form solution 
to calculate Shapley value, 
instead of defining the utility function as the accuracy of the pair-wise prediction, 
we define the utility function as follows: 

\begin{equation} \label{eq:utilitySingleDev}
    v(S) = 
    \left\{\vphantom{\begin{array}{c} a\\[6ex]\end{array}}\right.
        \begin{array}{ll}
            \hat{y}^{dev}_{p} - \hat{y}^{dev}_{q}, & {\rm if} \quad z^{dev}_{p,q} = 1, \\
            \hat{y}^{dev}_{q} - \hat{y}^{dev}_{q}, & {\rm if} \quad z^{dev}_{p,q} = 0
       \end{array}
\end{equation}

\noindent \textbf{Theorem 1} 
\textit{Consider the utility function in Equation (\ref{eq:utilitySingleDev}). Then the Shapley value of each training point $s_m$ can 
be decomposed into two terms $s^{(p)}_{m}$ and $s^{(q)}_{m}$ which depend on 
$x^{dev}_{p}$ and $x^{dev}_{q}$ respectively.  
$s^{(p)}_{m}$ and $s^{(q)}_{m}$ can 
be calculated 
recursively as follows: 
}

\begin{align*}
    s_m & =     
        \left\{\vphantom{\begin{array}{c} a\\[6ex]\end{array}}\right.
        \begin{array}{ll}
            s^{(p)}_{m} - s^{(q)}_{m}, & {\rm if} \quad z^{dev}_{p,q} = 1, \\
            s^{(q)}_{m} - s^{(p)}_{m}, & {\rm if} \quad z^{dev}_{p,q} = 0 \\
       \end{array} \\
    s^{(p)}_{\alpha^{(p)}_{N}}  &= \frac{ y_{\alpha^{(p)}_{N}} }{N} \qquad \qquad 
    s^{(q)}_{\alpha^{(q)}_{N}} = \frac{ y_{\alpha^{(q)}_{N}} }{N} \\
    s^{(p)}_{\alpha_{m}} & = s^{(p)}_{ \alpha^{(p)}_{m+1}} + 
    \frac{ y_{\alpha^{(p)}_{m}} - y_{\alpha^{(p)}_{m+1}} }{K} 
    \frac{min\{K, m\}}{m} \\ 
    s^{(q)}_{\alpha^{(q)}_{m}} & = s^{(q)}_{ \alpha^{(q)}_{m+1}} + 
    \frac{ y_{\alpha^{(q)}_{m}} - y_{\alpha^{(q)}_{m+1}} }{K} 
    \frac{min\{K, m\}}{m}
\end{align*}

With Theorem 1, 
the Shapley value calculation could be finished in $\mathcal{O}(N \log N)$ time. 
The above result for a single point in the development set 
could be readily extended to 
the multiple-testpoint case. 
In our experiment, with such optimization, 
the Shapley value calculation takes less than 5 seconds to finish. 
Theorem 1 comes primarily from~\cite{boxinShapley,boxinShapley2} and we extends 
their results of vanilla KNN regressor~\cite{boxinShapley} to our pairwise testing setting.

By applying the Shapley technique to the data,  
we identify noisy training data points which contribute negatively to the performance and remove them from the training set. 
Similar to Stage 2,
to construct training dialog pairs, 
we randomly sample dialog pairs $x_i$ and $x_j$ 
from the cleaned training set 
and derive 
$ z_{i,j} $ by comparing the self-reported rating ${y_i}$ and ${y_j}$. 
We then further fine tune the model from Stage 2. 
Theoretically, 
we could iterate between Stage 2 and Stage 3 multiple times 
while in practice one iteration is enough.

\subsection{Towards Scalable Pair-based Training}
We use a similar factorization technique for pair-wise ranking in  LambdaRank~\cite{lambdarank} to speed up training. 
For Stage 2 and 3, we have $\mathcal{O}(N^2)$ possible dialog pairs, which leads to quadratically increasing training time. 
Similar to LambdaRank~\cite{lambdarank}, 
it is possible to calculate the exact gradient of $\mathcal{O}(N^2)$ possible dialog pairs 
with $\mathcal{O}(N)$ forwards and back-propagations. 
More specifically, 
we denote the possible input pairs during training at Stage 2 or Stage 3 as: 
$ D^{pair}_{train} = 
\{(x_i, x_j, z_{i,j})\}_{i,j \in I}$. 
The total cost $L$ for $\mathcal{O}(N^2)$ possible dialog pairs 
is the sum of $\mathcal{O}(N^2)$ cross-entropy costs: 
\begin{align*}
    L_{i,j} & =  CrossEntropy(\hat{z}_{i,j}, {z_{i,j}})\\ 
    L & = \sum_{(i,j) \in I } L_{i,j}
\end{align*}

\noindent \textbf{Theorem 2} 
\textit{We can compute   $\frac{ \partial L  }{ \partial w_k } $ in $\mathcal{O}(N)$ by factor it  
into a weighted sum of $\frac{ \partial o_i }{ \partial w_k}$ where the weight $ \lambda_{i} \in  \mathbb{R}$ only depends on 
$\{ o_j\}$ and $\{z_{i,j}\}$. W.l.o.g., we assume $z_{i,j} \equiv 1$ . 
}
\begin{equation*}
    \frac{ \partial L  }{ \partial w_k } 
    = \sum_{i} 
    \lambda_{i}
    \frac{ \partial o_i }{ \partial w_k  }
\end{equation*}
\textit{and}
\begin{equation*} \label{eq:lambda}
    \lambda_{i} 
    = \sum_{j:(i,j) \in I} \frac{-1}{1+e^{o_i-o_j}} 
    + \sum_{j:(j,i) \in I} \frac{1}{1+e^{o_i-o_j}} \\
\end{equation*}

Here $ o_i = f(\phi(x_i)) \in \mathbb{R}$ and $ o_j = f(\phi(x_j)) \in \mathbb{R}$ are the outputs of the two branches of the model.
Theorem 2 shows that instead of performing back-propagation for all possible pairs, we could first perform $N$ forward passes to obtain $\{o_j\}$ and then calculate $\{\lambda_{i}\}$. 
Calculating $\{\lambda_{i}\}$ from $\{ o_j\}$ 
in Equation~\ref{eq:lambda} 
takes negligible time since this stage does not involve any neural network operation. Finally, we calculate a weighted sum of $\mathcal{O}(N)$ back-propagation and update the model parameters.

\section{Experiment}
\label{sec:experiment}

\paragraph{Model Setup}
We fine tune the pre-trained BERT~\cite{bert,DBLP:journals/corr/abs-1912-10375,liang2020bond} to learn the dialog feature extractor $\phi$. We partition the 403 expert annotated dialog pairs into a 200-pair development set and a 203-pair test set. 
We set $K=50$ for both the KNN label smoothing in Stage 2 and the KNN Shapley value calculation in Stage 3.

\paragraph{Model Details}
The details of extending BERT to encode multi-turn dialogs are as follows.  
Each dialog is represented as a sequence of tokens 
in the following input format: 
Starting with a special starting token $[CLS]$, 
we concatenate tokenized user and system utterances in chronological order with $[SEP]$ as the separators for adjacent utterance. 
In other words, we represent each dialog as a sequence: $[CLS]$, $S_{1,1}$, $S_{1,2}$, $...$, $[SEP]$, $U_{1,1}$, $U_{1,2}$, $...$, $[SEP]$, $S_{2,1}$, $S_{2,2}$, $...$, $[SEP]$ 
where $S_{i,j}$ and $U_{i,j}$ are the $j^{th}$ token 
of the system and user utterance in the $i^{th}$ turn. 
Following BERT, 
we also add a learned embedding to 
every token indicating whether it comes from user utterances or system utterances
. 

\begin{table}[htb]
\small
\begin{center}
\begin{tabular}{lrlll}
\cmidrule[\heavyrulewidth]{1-5}
\multirow{2}{*}{\bf No.} 
& \multirow{2}{*}{\bf Model} 
& \multirow{2}{*}{\bf \begin{tabular}[c]{@{}c@{}} Test \\ Acc. 
\end{tabular}}  
&\multicolumn{2}{l}{\bf Kappa} 
\\
\cmidrule{4-5}
 & & & $\kappa$ & $SE$ \\
\cmidrule{1-5}
(1) & BERT-Classification & 0.581 &0.161 & 0.049\\
(2) & BERT-Regression & 0.640 & 0.280 & 0.048\\
(3) & BERT-Pairwise & 0.730 & 0.459 & 0.044\\
(4) & BERT-Pairwise+Dev & 0.749 & 0.499 & 0.043\\ 
\cmidrule{1-5} 
(5) & Stage 2 & 0.755 & 0.509 & 0.043 \\ 
(6) & Stage 2 + 3 & 0.764 & 0.529 & 0.042 \\
\cmidrule{1-5}
(7) & Stage 3 & 0.714 & 0.429 & 0.045 \\
(8) & Stage 1 & 0.620 & 0.241 & 0.048 \\ 
(9) & Stage 1 + 3 & 0.788 & 0.628 & 0.039 \\ 
(10) & Stage 1 + 2 & 0.837 & 0.673 & 0.037 \\ 
(11) & \bf \modelabbrevname{} 
& \bf 0.892 & \bf 0.787 & \bf 0.031 \\ 
\cmidrule[\heavyrulewidth]{1-5}
\end{tabular}
\caption{
Test accuracy and kappa agreement comparison among variants of \modelabbrevname{}.
}
\label{tab:mainresult}
\end{center}
\vspace{-5mm}
\end{table}

\paragraph{Model Comparisons and Ablations}
We compare \modelabbrevname{} to its several ablations (Table~\ref{tab:mainresult}) and evaluate the performance 
on the testing set, which is annotated by experts. 
We also report the kappa agreement ~\cite{cohen1968weightedkappa} 
(kappa $\kappa$ and Standard Error $SE$)
between the predicted output and the expert annotations. 
(1) BERT-Classification and (2) BERT-Regression 
fine tune the pre-trained BERT to perform a 5-class classification and regression respectively directly using the noisy self-reported ratings. 
To test BERT-Classification on dialog pairs,
we apply the DEX trick~\cite{ageEstimationDEX} to 
get a  
floating-point 
number of predicted rating 
and thus get rid of the cases when the model predicts 
the dialog pairs as tie. 
(3) BERT-Pairwise shares the same model architecture with \modelabbrevname. 
It constructs dialog pairs for training by randomly sample dialog pairs $x_i$ and $x_j$ and derive 
$ z_{i,j} $ by comparing the corresponding self-reported user rating ${y_i}$ and ${y_j}$. We discard the pairs with equal ${y_i}$ and ${y_j}$. 
(4) BERT-Pairwise+Dev augments (3) 
by adding the 200 expert annotated dialog pairs in the development into the training data. 
We also compare the variants of 
\modelabbrevname{} which 
skips one or two of the three stages. 

\paragraph{Results} 
 Our first takeaway is that vanilla classification or regression formulation might not be the best way to formulate the problem of learning a dialog evaluation model. 
As shown in Table~\ref{tab:mainresult}, 
pairwise architecture (BERT-Pairwise, 0.73) is better than classification (BERT-Classification, 0.53) or regression (BERT-Regression, 0.64) in this problem. 
Similar to our observation, 
the research community in computer vision 
has long observed that both vanilla classification and regression 
formulation has drawbacks in age  estimation~\cite{ageEstimationDEX,ageEstimationOrdinalRegression,ageEstimationPairPosterior}.

Our second takeaway is that 
denoising algorithm that is more aggressive 
usually makes stronger assumptions on the quality of feature representations. 
Therefore, it helps to create a denoising pipeline that starts with 
better feature representation learning and 
less aggressive denoising algorithm 
to learn better feature representation before applying the more aggressive denoising algorithms. 
As shown in Table~\ref{tab:mainresult}, 
our three-stage denoising pipeline 
\modelabbrevname{} (Acc. 0.892) significantly outperforms all baselines by a large margin. 
Although (8) Stage 1 does not directly provide high accuracy (Acc. 0.620), the feature representation it learned is extremely important. 
Without Stage 1, both (5) Stage 2 (Acc. 0.755) and (6) Stage 2 + Stage 3 (Acc. 0.763) perform worse. 

Since the KNN label smoothing is performed on the feature space, we expect the smoothing performs worse 
without self-supervised dialog feature representation learning in Stage 1. 
However, they still work better than baseline (1) (2) (3) which are models that do not 
account for the noise in data. 
This is because we use the pre-trained BERT to initialize our dialog encoder $\phi$ and thus $\phi$ is still able to provide some useful features for Stage 2. 
In addition, we observe that denoising with data Shapley in Stage 3 
requires better dialog feature representation. 
(7) Stage 3 (Acc. 0.714) performs even worse than BERT-Pairwise (0.730) without good representations to perform the Shapley denoising algorithm. 
Skipping Stage 2 also hurts performance (Acc. 0.788). However, it does not mean that Shapley denoising in Stage 3 is not powerful. 
We observe a large performance gain in applying stage 3 after stage 1 and stage 2 (Acc. 0.837 v.s. 0.892). 
Finally, we note that adding the expert annotated development set directly into the training data is much less efficient compared to using the development set for data Shapley to denoise. 
BERT-Pairwise+Dev only reaches an accuracy of 0.749.


\begin{table}[hb]
\vspace{-3mm}
\footnotesize
\centering
\begin{tabular}{p{7cm}}
\cmidrule[\heavyrulewidth]{1-1}
\textbf{User:}  the whole harry potter series  \\
\textbf{Sys:}  whoops i think i misheard you.  can you repeat the movie name please?  \\
\textbf{User:}  all the harry potter series  \\
\textbf{Sys:}  ah, i don't know any movies by that name.  sorry if i sometimes confuse you saying something else with a movie name ...  \\ 
\multicolumn{1}{c}{...} \\ 
\textbf{Sys:}  oh, okay.  anyways, i really enjoyed watching black panther and wonder woman recently.  what's your favorite superhero movie?  \\
\textbf{User:}  super man  \\
\textbf{Sys:}  what was that?  can you please say the movie name again?  \\
\textbf{User:}  stop i gotta go  \\
\cmidrule[\heavyrulewidth]{1-1}
\end{tabular}
\caption{
An example dialog with noisy self-reported rating (Score-5)  identified by data shapley. 
}
\label{table:exampleDialogNegShapleyValue}
\vspace{-5mm}
\end{table}

\begin{figure}[t]
\centering
\includegraphics[width=0.45\textwidth]{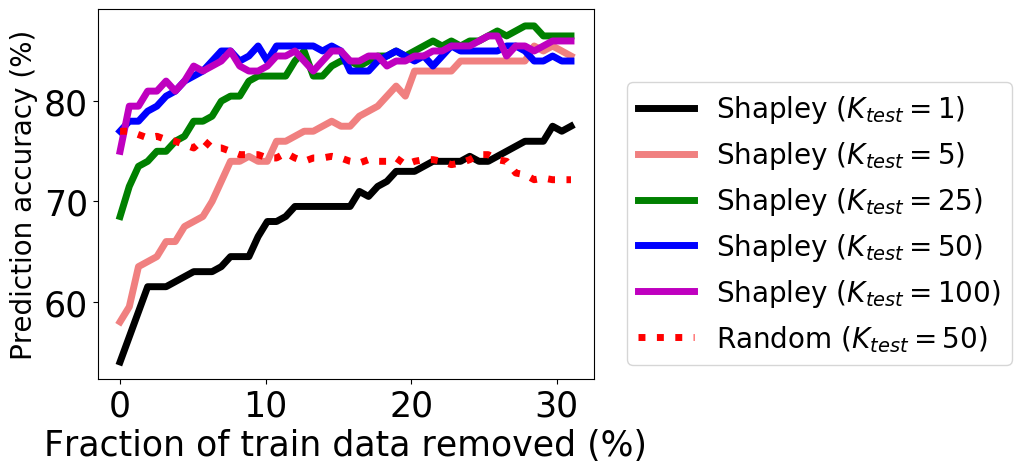}
\caption{
Removing training data with low Shapley value improves the performance of the KNN regressor. 
}
\label{fig:shapleyremove1}
\vspace{-5mm}
\end{figure}

\paragraph{Additional analysis}
We also present an analysis to show how Shapley denoising works as shown in Figure~\ref{fig:shapleyremove1}. 
We examine the Shapley value for each training datum in Stage 3. 
We first show an example dialog with a negative Shapley value in Table~\ref{table:exampleDialogNegShapleyValue}. 
According to the Shapley value, 
we remove data points one by one starting from the least valuable to the most valuable. 
Each time, after the point is removed, 
we create new KNN regressor models $K=1,5,25,50,100$ on the remaining dialogs and labels and evaluate them on the test set with expert annotations. 
We extract the features of the remaining dialogs using the dialog encoder $\phi$ tuned by Stage 1 and Stage 2. 
As shown in Figure~\ref{fig:shapleyremove1}, 
removing training data with low Shapley values increases the performance to a certain point before convergence for $K$ of all choices. We observe a similar trend when re-training a model on the remaining data. 
In contrast, removing data randomly decreases the performance on the test set. 
In addition, larger $K$ has a better performance, which validates the denoising effect of KNN with large $K$.

\section{Conclusion}
The ultimate chatbot evaluation metric should be user-centric, as chatbots are there to provide human with an enjoyable experiences. 
Previously Likert-score based self-reported rating is the de-facto standard 
for current dialog evaluation . 
However, our analysis indicates that self-reported dialog ratings are skewed (J-shape), noisy and insensitive due to bias and variance among different users. We propose a three-stage denoising pipeline \modelabbrevname{} 
to reduce self-reported ratings and, at the same time, build an automatic comparison-based automatic dialog quality predictor.
\modelabbrevname's results 
highly correlate with expert judgments on pair-wise dialog comparison ratings (89.2\% agreement, 0.787 Kappa). 

\section*{Acknowledgments}
We would like to sincerely thank ACL 2020 Chairs and Reviewers for their review efforts and helpful feedback. 
We thank Yu Li for his insightful guidance and support in shaping this project. 
We thank Boxin Wang for helpful discussions on data Shapley. 
We would also like to extend our gratitude to Yanbang Wang, Youzhi Tian, Weiyan Shi, Allen Nie and Michihiro Yasunaga for their valuable feedback and suggestions.

\bibliographystyle{acl_natbib}
\bibliography{acl2020.bib}

\end{document}